\documentclass[10pt,twocolumn,letterpaper]{article}

\usepackage{iccv}
\usepackage{times}
\usepackage{epsfig}
\usepackage{rotating,graphicx}
\usepackage{amsmath}
\usepackage{amssymb}
\usepackage{dblfloatfix}
\usepackage{algorithm}
\usepackage{fixltx2e}
\usepackage{mathtools}
\usepackage[noend]{algpseudocode}
\usepackage{supertabular,booktabs}
\usepackage{subcaption}
\usepackage{placeins}
\usepackage{flafter}
\usepackage{enumitem}
\usepackage{multirow}
\usepackage{booktabs}
\usepackage{soul}
  \usepackage{longtable}
\usepackage[pagebackref=true,breaklinks=true,letterpaper=true,colorlinks,bookmarks=false]{hyperref}

 \iccvfinalcopy % *** Uncomment this line for the final submission

\def\iccvPaperID{3565} % *** Enter the ICCV Paper ID here

\ificcvfinal\fi
\begin{document}

%%%%%%%%% TITLE
\title{A New Stereo Benchmarking Dataset for Satellite Images}

\author{Sonali Patil 
\hspace{1.5cm} Bharath Comandur 
\hspace{1.5cm} Tanmay Prakash 
\hspace{1.5cm} Avinash C. Kak \\
School of Electrical and Computer Engineering, Purdue University\\
West Lafayette, IN, USA\\
            {\tt\small patil19@purdue.edu} 
\hspace{1.5cm}{\tt\small bcomandu@purdue.edu}
\hspace{1.5cm}{\tt\small tprakash@purdue.edu}
\hspace{1.5cm}{\tt\small kak@purdue.edu}
}

%% omit the following lines up until the closing ``}''.
%% Additional authors and addresses can be added with ``\and'',
%% just like the second author.
%% To save space, use either the email address or home page, not both
%\and

\maketitle
%\thispagestyle{empty}

%%%%%%%%% ABSTRACT
\begin{abstract}
In order to facilitate further research in stereo
reconstruction with multi-date satellite images, the goal of
this paper is to provide a set of stereo-rectified images
and the associated groundtruthed disparities for 10 AOIs
(Area of Interest) drawn from two sources: 8 AOIs from
IARPA's MVS Challenge dataset and 2 AOIs from the
CORE3D-Public dataset.  The disparities were groundtruthed
by first constructing a fused DSM from the stereo pairs and
by aligning 30 cm LiDAR with the fused DSM.  Unlike the
existing benckmarking datasets, we have also carried out a
quantitative evaluation of our groundtruthed disparities
using human annotated points in two of the AOIs.
Additionally, the rectification accuracy in our dataset is
comparable to the same in the existing state-of-the-art
stereo datasets.  In general, we have used the WorldView-3
(WV3) images for the dataset, the exception being the UCSD
area for which we have used both WV3 and WorldView-2 (WV2)
images.  All of the dataset images are now in the public
domain.  Since multi-date satellite images frequently
include images acquired in different seasons (which creates
challenges in finding corresponding pairs of pixels for
stereo), our dataset also includes for each image a building
mask over which the disparities estimated by stereo should
prove reliable.  Additional metadata included in the dataset
includes information about each image's acquisition date and
time, the azimuth and elevation angles of the camera, and
the intersection angles for the two views in a stereo pair.
Also included in the dataset are both quantitative and
qualitative analyses of the accuracy of the groundtruthed
disparity maps. Our dataset is available for download at 
\url{https://engineering.purdue.edu/RVL/Database/SatStereo/index.html}
\end{abstract}

%%%%%%%%% BODY TEXT

\section{Introduction}
\label{sec:intro}
While there now exist several datasets for projective
cameras that can be used to test the performance of stereo
matching algorithms, the same cannot be said for the
pushbroom cameras used for satellite images.  The images
produced by the pushbroom cameras are particularly
challenging for stereo disparity calculations because their
epipolar lines (which in reality are curves) do not form
conjugate pairs.  Additional sources of difficulty are the
facts that the images are generally recorded at different
times and during different seasons, which creates
difficulties in establishing matches between 
%what would otherwise be 
the corresponding pixels in the images of a stereo pair.

\begin{figure}[h]
  \begin{center}    
\includegraphics[width=\linewidth]{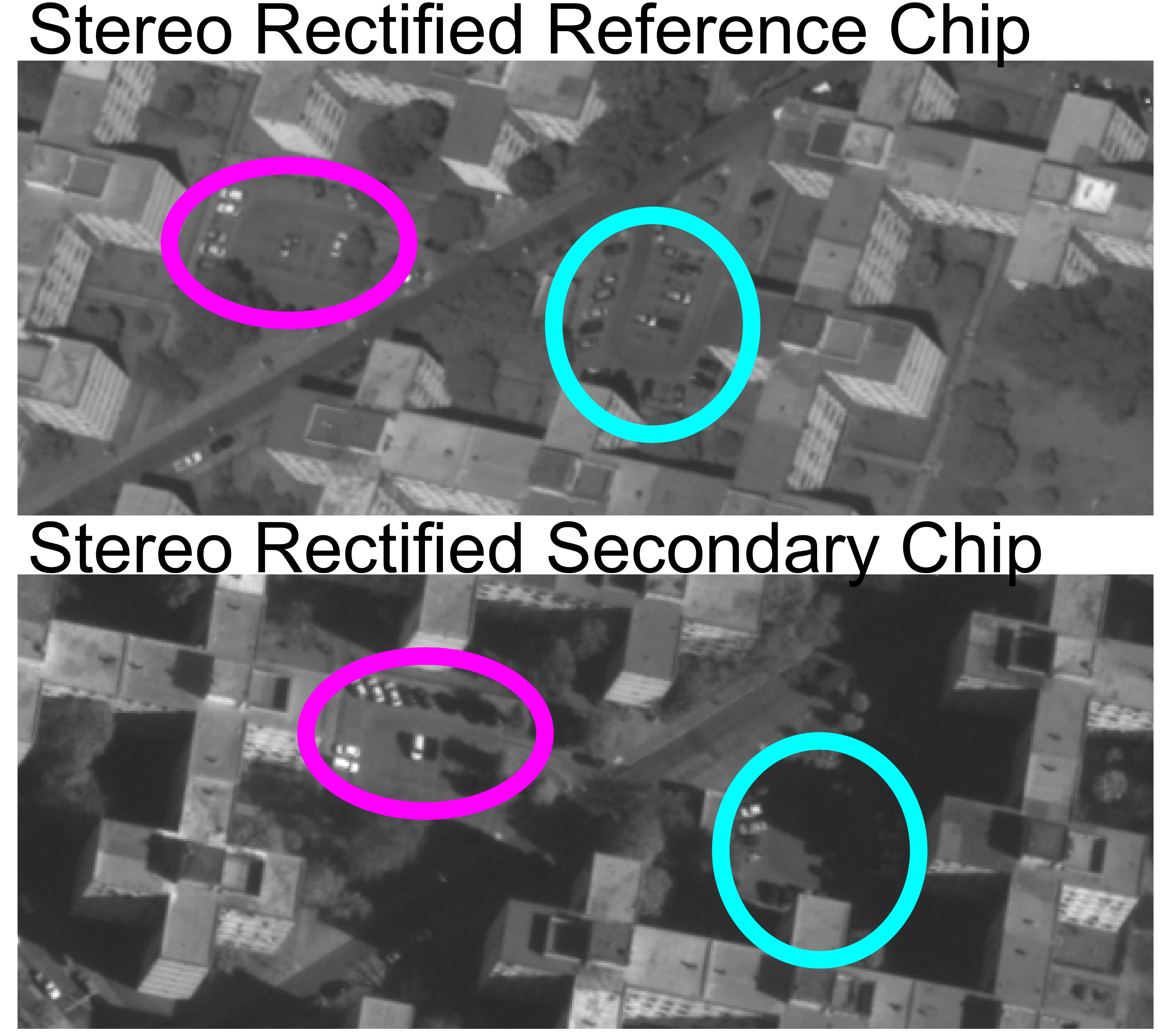}
\end{center}
   \caption{A stereo rectified chip pair from the MP2 AOI,
     with the images acquired roughly six months apart.
     This example illustrates significant variations in the
     ground-level scene content.  The highlighted
     ground-level regions show changes in the parking lots
     due to changes in the number of cars and the changes in
     the shadows.}
\label{fig:ground}
\end{figure}

The goal of this paper is to remedy this shortcoming in the
research community by providing a dataset with groundtruthed
disparities for: (1) eight AOIs (Area of Interests) in San
Fernando, Argentina, as defined in the IARPA's MVS Challenge
dataset \cite{MVS_Challenge2016}.  That challenge dataset,
spanning 100 square kilometers, includes 30 cm airborne
LiDAR groundtruth data for a 20 square kilometer subset of
the larger area. All this data is now publicly available and
we have utilized it for constructing our groundtruthed
disparity maps. And (2) two AOIs from the CORE3D-Public
dataset \cite{Core3D_Public} covering areas in UCSD and
Jacksonville.  In general, we have drawn stereo pairs from
WV3.  The exception is the UCSD area, for which we have used
both WV2 and WV3 images.
%%%%%%%%%%%%%%%%%%%%%%%%%
\begin{figure*}
\begin{center}
%\fbox{\rule{0pt}{2.5in} \rule{.9\linewidth}{0pt}}
\includegraphics[width=\textwidth]{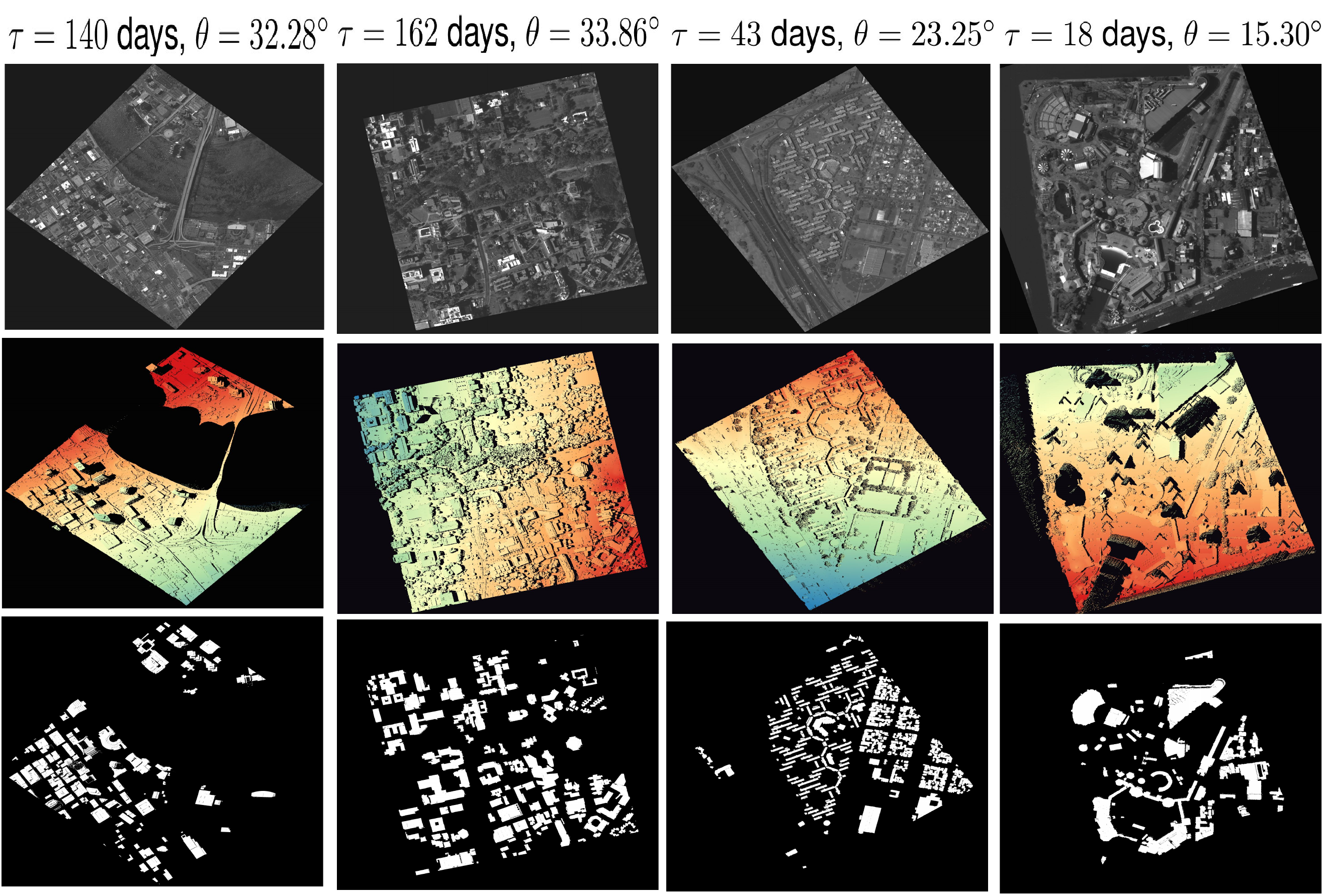}
\end{center}
   \caption{Qualitative results for our groundtruthed
     disparity maps from the top four largest AOIs. From
     left to right the AOIs are: Jacksonville, UCSD WV2,
     Explorer, and MasterSequestered (MS) Park. For
     additional results, please see the supplemental
     material submitted with this manuscript. Note that the
     large holes in MS Park are either in the water regions
     or the surrounding land regions. Since the LiDAR output
     is sensitive to specularly reflective surfaces, the
     heights as provided by LiDAR in such regions are either
     invalid values or large noise spikes. $\tau$ and $\theta$ represent the time difference between a stereo pair and the intersection angle, respectively.}
\label{fig:qualitative_results}
\end{figure*}
%%%%%%%%%%%%%%%%%%%%%%%%%
The AOIs included in the dataset vary in size from 0.1
sq. km.  to 2 sq. km. and the number of selected stereo pairs for the
different AOIs varies from 53 to 505.  One of the major
challenges presented by out-of-date satellite imagery is
that some of the scene content may vary from image to image.
Figure \ref{fig:ground} illustrates how in the MVS MasterProvisional 2
area variations in scene content are more prominent in
ground regions as compared to building regions.  Therefore,
we provide building masks to mark regions where we can
expect relatively high accuracies for the groundtruthed
disparities.  This is not to imply that the groundtruthed
disparities for the ground regions are always unreliable.

In order to facilitate research in large-scale stereo
reconstruction, we also provide additional metadata for each
stereo pair, which includes the information about each
image's acquisition date and time, the azimuth and elevation
angles for the images, and the intersection angle of the two
views.  Such metadata can help to see a correlation between
these parameters and stereo matching quality.

A unique aspect of our dataset is that we have carried out a
quantitative evaluation of our groundtruthed disparities
using human annotated points in two AOIs. Additionally, we
have evaluated the stereo rectification accuracies; the
rectification errors are less than 0.5 pixels on the
average.

The dataset itself consists of a set of selected
stereo-rectified image pairs for each of the ten AOIs and
presents the groundtruthed disparities for each pixel in a
reference image and the corresponding secondary image in
each pair.  The stereo pair selection is based on time
difference and view-angle difference considerations.  For
each AOI, the disparities are groundtruthed by first
constructing a fused DSM (Digital Surface Model) from the
stereo pairs and, subsequently, by aligning the LiDAR with
the fused DSM.  This process of aligning LiDAR with the
fused DSM allows for a mapping from each pixel (x,y) in the
reference rectified image of a stereo pair to what the true
value of the disparity should be at that pixel.  The process
of extracting the groundtruthed disparities at the different
pixels in a reference stereo-rectified image involves back
projecting pixels of the rectified image and using the
lat/long coordinates thus calculated to get the
corresponding height from the LiDAR data.  Subsequently,
this height value is translated into the disparity value.
Also included in the dataset are both quantitative and
qualitative analyses of the accuracy of the groundtruthed
disparity maps.

Table \ref{tbl:data_summary} presents the summary of our
dataset. Figure \ref{fig:qualitative_results} shows some
examples of our disparity maps from the top four largest AOIs,
along with their corresponding building masks and metadata.
%Our dataset is available for download at 
%\url{https://engineering.purdue.edu/RVL/Database/SatStereo/index.html}

%\textcolor{red}{\Large\bf Sonali, please explain MS Park artifacts.}

\begin{table}[h]
\centering
% \footnotesize \small
{\small
\begin{tabular}{p{2.2cm}|p{1.0cm}|p{1.0cm}|p{1.0cm}}
\toprule
 Dataset & WV3 / WV2& Area (sq. km. )&  No. of Selected Stereo Pairs   \\
\hline
MP 1 (MVS) & WV3 & 0.13 & 505 \\
MP 2 (MVS) & WV3 & 0.14 & 361 \\
MP 3 (MVS) & WV3 & 0.10  & 246 \\
Explorer (MVS) & WV3 & 0.45 & 329 \\
MS 1 (MVS) & WV3 & 0.12 & 501 \\
MS 2 (MVS) & WV3 & 0.13 & 499 \\
MS 3 (MVS) & WV3 & 0.12 & 349  \\
MS Park (MVS) & WV3 & 0.25  & 301  \\
UCSD (CORE3D) & WV2 & 1 & 336 \\
UCSD (CORE3D) & WV3 & 1 & 130 \\
%\multirow{2}{*}{UCSD} (CORE3D) & WV2 & 1 & 336 \\
%    &WV3 & 1 & 182 \\
Jax (CORE3D) & WV3 & 2 & 53 \\
\bottomrule
\end{tabular} }
\caption{MVS stands for IARPA's Multi-View Stereo Challenge, MP stands
  for MasterProvisional and MS stands for MasterSequestered}
\label{tbl:data_summary}
\end{table} 

The rest of the paper is organized as follows: Section 2
briefly outlines the related work. In Section 3 we explain
how we use the notions of ``chips'' and ``tiles'' as used in
groundtruthing the disparity maps. We take up image-to-image
and LiDAR-to-fused DSM alignment issues in Section
4. Section 5 shows how aligned LiDAR is used to calculate
the groundtruth disparities. In Section 6 we present a
quantitative evaluation of the groundtruth disparity maps
and some benchmarking results of existing stereo matching
algorithms. Finally in Section 7 we conclude our findings.

%-------------------------------------------------------------------------
%\subsection{Language}
\section{Related Work}
\label{sec:related_work}
3D reconstruction is a popular area of research in the
computer vision community and there exist a number of
groundtruthed datasets for benchmarking stereo matching
algorithms. Although synthetic datasets created using
rendered scenes such as the MPI Sintel stereo dataset
\cite{Butler:ECCV:2012} might prove useful for certain
tasks, they do not necessarily capture the diversity and
complexity of images of the real world. Since our dataset
has been created to serve as a benchmark for binocular
stereo, it is sufficient to restrict our discussion of
related work to datasets that focus on binocular stereo. The
well known Tsukuba image pair \cite{nakamura1996occlusion}
was one of the first stereo datasets and contains disparity
maps created using manual annotation. Since then, multiple
attempts have been made to create more accurate datasets and
some of the most popular ones include the Middlebury, KITTI
and ETH3D datasets.

The Middlebury datasets include the Middlebury2001
\cite{Middlebury2001}, Middlebury2003 \cite{Middlebury2003},
Middlebury2005 and Middlebury2006 datasets
\cite{Middlebury5or6} and more recently the high resolution
Middlebury 2014 dataset \cite{Middlebury2014}. The last
dataset was created using a stereo rig with cameras and
structured light projectors and claims subpixel-accurate
groundtruth. Images are of resolution (5-6MP) and mostly
contain indoor scenes. Pairs are grouped under different
categories such as similar and varying ambient illumination,
perfect and imperfect rectification etc. Note that less than
50\% of the scenes required manual cleanup
\cite{Middlebury2014}.

With a focus on autonomous driving, the KITTI2012
\cite{KITTI2012} and KITTI2015 \cite{KITTI2015} datasets
were created to capture outdoor scenes. While the former
pays attention to static environments, the latter is
concerned with moving objects captured by a stereo
camera. For generating groundtruth, scans were captured
using a laser scanner mounted on a car and scenes were
annotated using 3D CAD models for moving vehicles. The
disparity maps in this dataset are semi-dense when compared
to the Middlebury2014 dataset.

The ETH3D \cite{eth3d} dataset was created to address some
of the shortcomings of the the above mentioned datasets
including small size, lower diversity, absence of outdoor
scenes (Middlebury), low resolution and sparseness
(KITTI). Groundtruth depth was captured using a high
precision laser scanner. It covers both indoor and outdoor
scenes and can be used to evaluate both binocular and
multi-view stereo algorithms.

The datasets described thus far consist of images taken with
projective cameras that are either handheld or mounted on
stereo rigs. Recently, there was an announcement of a stereo
dataset for satellite images \cite{U3D} that also provides
groundtruthed disparities. That dataset however does not
provide estimates of the errors in the groundtruthed
disparities using human annotated points. Additionally, that
dataset also does not present any information on the
rectification errors involved. Note also that the framework
we have used for creating the dataset is significantly
different from the one used in \cite{U3D}. We believe that
the research community can only benefit by experimenting
with datasets produced with two different approaches.

% To
% the best of our knowledge, there does not exist a
% benchmarking dataset for images taken using pushbroom
% cameras from satellites orbiting at great speeds (thousands
% of meters/sec) at high altitudes (600 - 800 km) above the
% Earth's surface. The complexities and challenges presented
% by the data collection process and sensor geometry
% notwithstanding, the highly diverse nature of the Earth's
% surface and atmospheric conditions introduces variations in
% illumination, scene content, texture, spectral signature
% etc. at significantly greater scales than exhibited in any
% of the aforementioned datasets.
% Furthermore, the nature of
% the data makes generation of groundtruth disparity maps that
% much more challenging. Our dataset attempts to fill this
% vacuum for accurate stereo benchmarking datasets for such
% remote sensing sensors. We hope that researchers working on
% the complicated problem of 3D reconstruction from satellite
% images will find our dataset useful for developing and
% evaluating their stereo matching algorithms.

%\section{Data Alignment}

%\section{Chip versus Tile and the Overall Processing Architecture}

\section{Chip versus Tile Conundrum}
\label{sec:chip_vs_tile}
%for Generating Groundtruthed Disparities}

Each of the AOIs in our dataset is specified by a KML
polygon in the lat/long space.  The portions of the
satellite images extracted through each of these polygons
are referred to as {\em chips}.
%The KML-extracted section of each image for any given AOI
%will be referred to as a chip.  
Chips come in varying sizes, depending on the size of the
AOI. The largest of the chips are of size roughly $5000
\times 5000$.  Our goal is to provide groundtruthed
disparities at the chip level so that a disparity map would
cover the entire AOI.  

%As mentioned earlier, the dataset we
%provide consists of pairs of stereo-rectified chips, along
%with the groundtruthed disparities at each pixel in the
%reference image for every pair of the chips extracted from a
%selected subset of all possible pairs of images over an AOI.

Unfortunately, on account of the fact that, in general, the
pushbroom cameras used in the satellites are characterized
by non-conjugate epipolar curves, it is possible for the
chip pairs to be much too large for a straightforward
implementation of stereo rectification.  There do exist two
different approaches to get around this difficulty: (1) To
use the approach suggested by Oh
et. al. \cite{oh2010piecewise} that consists of first
finding piecewise correspondences between the different
possible epipolar curve pairs and then resampling the
original images in a way that straightens out the epipolar
curves. And (2) To use the method proposed by Franchis et al
\cite{de2014automatic} that consists of breaking the chips
into smaller tiles under the assumption that the part of the
epipolar curve spanning a tile may be well approximated by a
straight line.  Another way of saying the same thing would
be that the image portions in the tiles may be considered to
have come from an affine imaging sensor.
Following \cite{de2014automatic}, we have used the latter approach.
%our image-to-image alignment pipeline is based on an
%extension of the s2p pipeline \cite{de2014automatic}.  The
The tiles that we use are typically of size $500 \times 500$.
%We directly use the s2p logic to break a chip into tiles.

That then takes us to the heart of the algorithmic problem
we needed to solve for generating the groundtruthed dataset:
How to jump from the initial stereo rectifications based on
tile based processing to the final stereo rectifications for
the chips that would be needed for the dataset?  While the
initial stereo rectification would be applied to the {\em
  tiles} that would typically be of size $500 \times 500$,
we would want to translate that into the final stereo
rectifications of the {\em chips} that may be as large as
$5000 \times 5000$ for the larger AOIs.

\begin{figure}[h]
\begin{center}
\includegraphics[width=\linewidth]{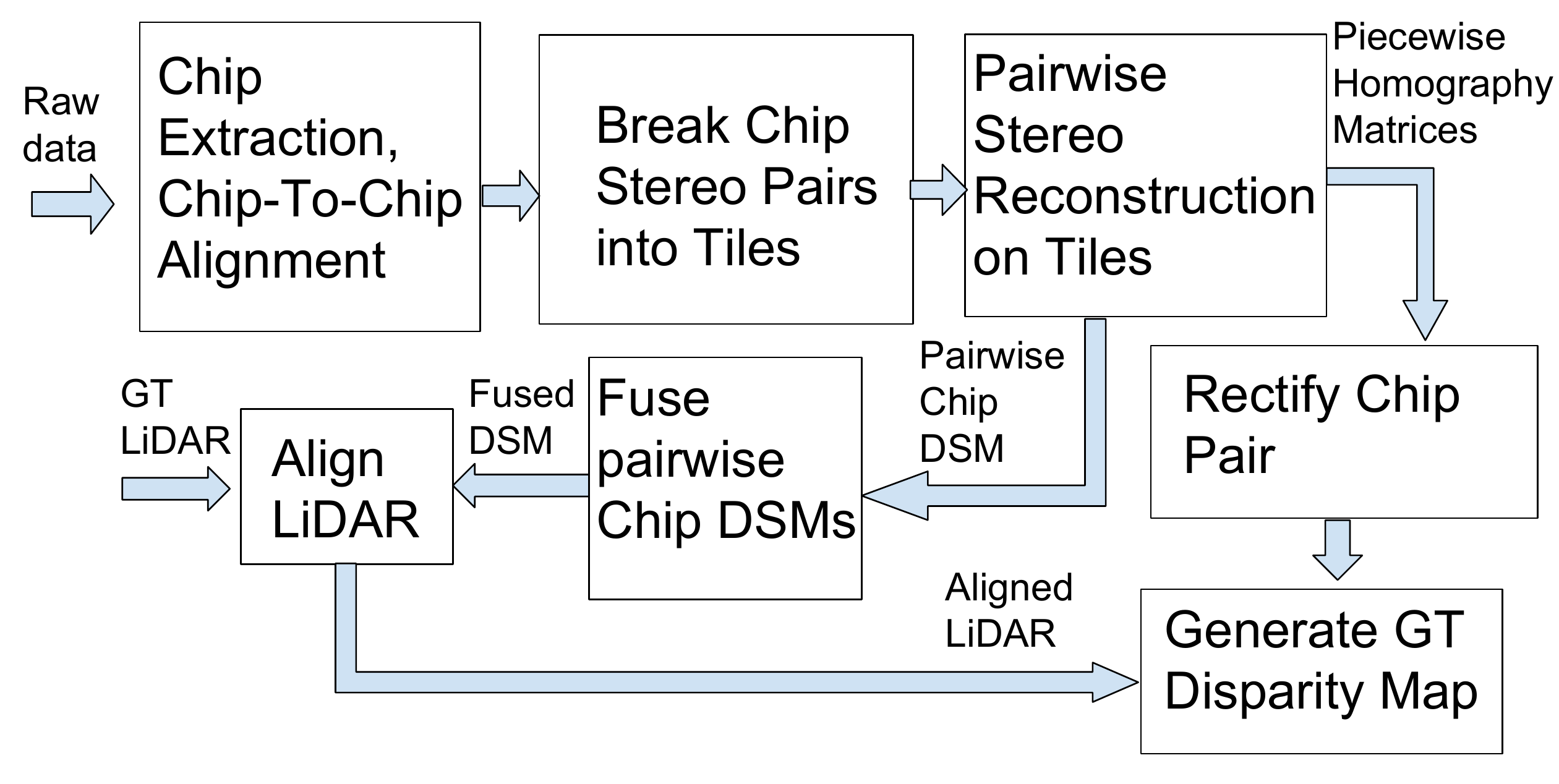}
\end{center}
   \caption{Overall pipeline of our approach, showing difference between chip-level processing and tile-level processing}
\label{fig:overall_pipeline}
\end{figure}

As to how we solve this chip vs. tile conundrum is best
explained through the overall processing architecture we
employ as shown in Fig. \ref{fig:overall_pipeline}.

As shown in the figure, the raw satellite images are first
subject to KML-based chip extraction for the AOIs.  All the
chips thus collected for each AOI are subject to pair
selection (not shown) followed by chip-to-chip alignment
and RPC bias correction as described in Section
\ref{sec:alignment} and its subsections.  As shown in the
middle box in the upper row, each chip is subsequently
broken into tiles; for this we use the s2p logic directly
\cite{de2014automatic}.  Breaking the chips into tiles
allows for conventional stereo rectification between pairs
of tiles, and that, in turn, allows for relatively easy
stereo reconstruction from the tiles. The tile-based stereo
reconstructions serve two purposes: (1) On a pairwise basis,
they yield the point clouds that when aggregated together
give us the chip-based pairwise DSMs as explained in Section
\ref{subsec:lidar_align}.  In Fig.
\ref{fig:overall_pipeline}, this is represented by the
downward pointing arrow that emanates from the box labeled
``Pairwise Stereo Rectification on Tiles''.  And (2), the
homographies that describe tile-based rectifications when modified
by tile locations inside the chips
%stereo reconstructions are also used for
%estimating the homographies that when modified by the
%location of a tile in a stereo rectified chip
give us the chip-based rectifications as explained in Section
\ref{subsec:chip_stereo_rect}.  The rest of Fig.
\ref{fig:overall_pipeline} should be self explanatory.

%\section{Image-to-Image Alignment and LiDAR-to-Fused DSM Alignment}
\section{Data Alignment}
\label{sec:alignment}

\begin{figure*}
\begin{center}
\includegraphics[width=\textwidth]{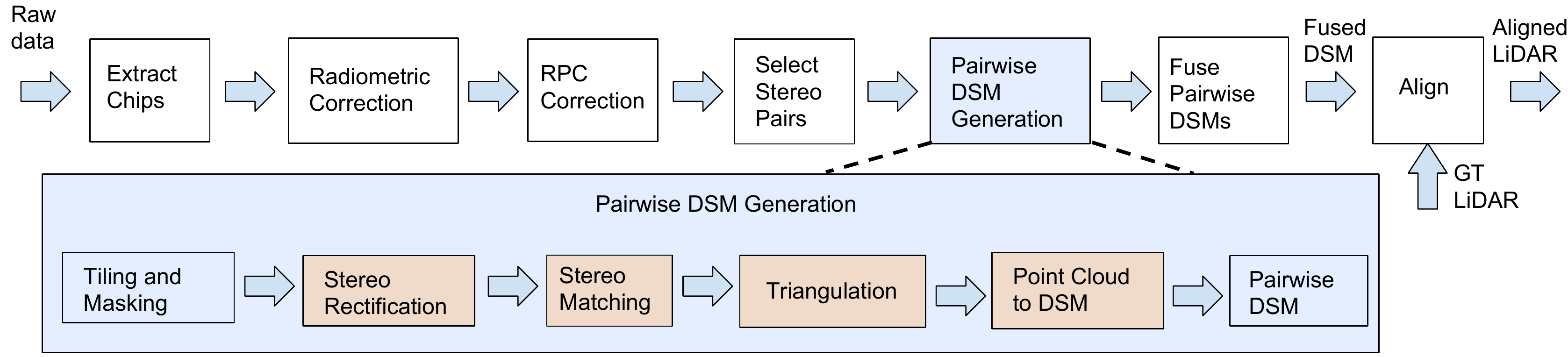}
\end{center}
\caption{The input raw data includes RPC files for camera
  parameters and NTF files for raw sensor data. SRTM
  (Shuttle Radar Tomography Mission) DEM (Digital Elevation
  Model) and KML vector which is, used to extract AOIs. The
  steps in orange color are processed in parallel per tile.}
\label{fig:data_alignment}
\end{figure*}

There are two types of data alignment carried out in the
processing architecture shown in Fig.
\ref{fig:overall_pipeline}: chip-to-chip alignment and the
LiDAR-to-fused-DSM alignment.  Both of these are captured by
the pipeline shown in Figure \ref{fig:data_alignment}.

%shows the data alignment we
%use for generating the dataset.  There are two parts to it:
%Note that both the chip-to-chip alignment and the
Note that the LiDAR-to-fused-DSM alignment is at the chip
level of processing.  Of the various steps shown in the
figure, we will cover chip extraction and radiometric
correction details in Section \ref{subsec:preprocess} that
is devoted to the preprocessing of the raw images.  That
will be followed in Section \ref{subsec:rpc_correct} by how
the RPC bias errors are estimated and the RPCs corrected.
Section \ref{subsec:pair_select} describes the logic we have
used for pair selection; this logic is based on a
combination of the difference in the image acquisition times
and the difference between the view-angles.  The other steps
shown in the pipeline are covered in Sections \ref{subsec:stereo_rect} and
\ref{subsec:lidar_align} .  

%---------------------------------
%%%%%%%%%%%%%%%%%%%%
\subsection{Preprocessing}
\label{subsec:preprocess}

%Each of the AOIs in our dataset is specified by a KML
%polygon in the lat/long space.  The portions of the
%satellite images extracted through each of these polygons
%are referred to as {\em chips}.

It is important to apply Top-of-Atmosphere correction when
using multi-date images. This involves the following steps --
1) converting the pixel values into ToA spectral radiance
i.e. the spectral radiance entering the telescope aperture
and subsequently 2) converting the ToA radiance to ToA
reflectance values, which effectively converts the earth-sun
distance to 1 Astronomical Unit (AU) and the solar zenith
angle to 0 degrees. Further details about the ToA correction
for WV2 and WV3 images can be found in \cite{TOA_WV2} and
\cite{TOA_WV3}, respectively.

% A standard practice in remote sensing, especially when using
% multi-view and multi-date data is to perform
% Top-of-Atmosphere (ToA) correction before any other
% processing. As a first step in ToA correction, using
% metadata supplied with the images, the pixel values measured
% by the sensor are converted to ToA spectral radiance
% i.e. the spectral radiance entering the telescope
% aperture. This ToA radiance for the same scene content on
% the ground will vary between the images as it depends on a
% number of factors, such as the solar zenith angle, the
% earth-sun distance, atmospheric effects etc. Using the
% metadata provided, we can convert the ToA radiance to ToA
% reflectance values, which effectively converts the earth-sun
% distance to 1 Astronomical Unit (AU) and the solar zenith
% angle to 0 degrees.  Further details about the ToA
% correction for WV2 and WV3 images can be found in
% \cite{TOA_WV2} and \cite{TOA_WV3}, respectively.

\subsection{RPC Correction}
\label{subsec:rpc_correct}
% Satellite image vendors provide initial camera models as a
% rational polynomial coefficient (RPC) model \cite{Dial2005}.
% The alignment of these initial camera models is sufficient
% for some purposes, such as backprojecting a pixel to within
% a few meters, but not for stereo matching.

According to \cite{Grodecki2003}, good alignment between
satellite images can be achieved by adding a constant bias
to the pixel locations output by the RPC model, rather than
explicitly updating the physical geometry of the camera. We
therefore correct the RPC model by jointly calculating the
appropriate biases for each RPC, using the popular approach
of bundle adjustment \cite{Triggs2000}.  % The RPC model is an
% approximation of the rigorous projection model (RPM) of the
% pushbroom sensor, but can generally match it to within a
% fraction of pixel.

Bundle adjustment aligns the images by jointly optimizing 3D
structure and the camera parameters over a global objective
function. First, we detect tie point correspondences between
images. Corresponding tie points are image points that have
been identified as the projections of the same 3D world
points. The reprojection error of a tie point is the
distance between a projection of an estimated world point
and the tie point. It is a function of both the camera
parameters and the world point.  By jointly finding the
world point coordinates and camera parameters that minimize
the total reprojection error over the full set of tie point
correspondences, we can align the images.

To populate the set of tie point correspondences, we compare
every possible pair of images. Given a pair of images, we
extract interest points in each image using SURF
\cite{Bay2006}, identify an initial set of tie point
correspondences between the two images using the SURF
feature descriptor, and prune outliers from that set of
correspondences using RANSAC.

\subsection {Stereo Pair Selection}
\label{subsec:pair_select}

Not all pairs --- especially so in the context of satellite
stereo reconstruction --- are equal.  Intuitively it makes
sense that disparity calculations would be aided by similar
scene content between images. Although designing a
theoretically correct way to select image pairs is a
difficult task, it is possible to use heuristics to improve
the chances of selecting good pairs. Along the lines of the
pair selection strategy used in
\cite{facciolo2017automatic}, we first apply thresholds to
the view angles to drop the highly off-nadir and the highly
near-nadir images.  We then select the pairs by applying
thresholds to the differences in the view angles and the
differences in times of acquisition between the images in
each pair. The pairs thus chosen are subsequently sorted in
increasing order of the time differences involved, the
intuition being that images captured closer in time have a
greater probability of having similar scene content. Note
that the WV2 and WV3 sensors are heliosynchronous, i.e. they
image the same location on the earth's surface at roughly
the same time every day.

\begin{figure*}
\begin{center}
\includegraphics[width=\textwidth]{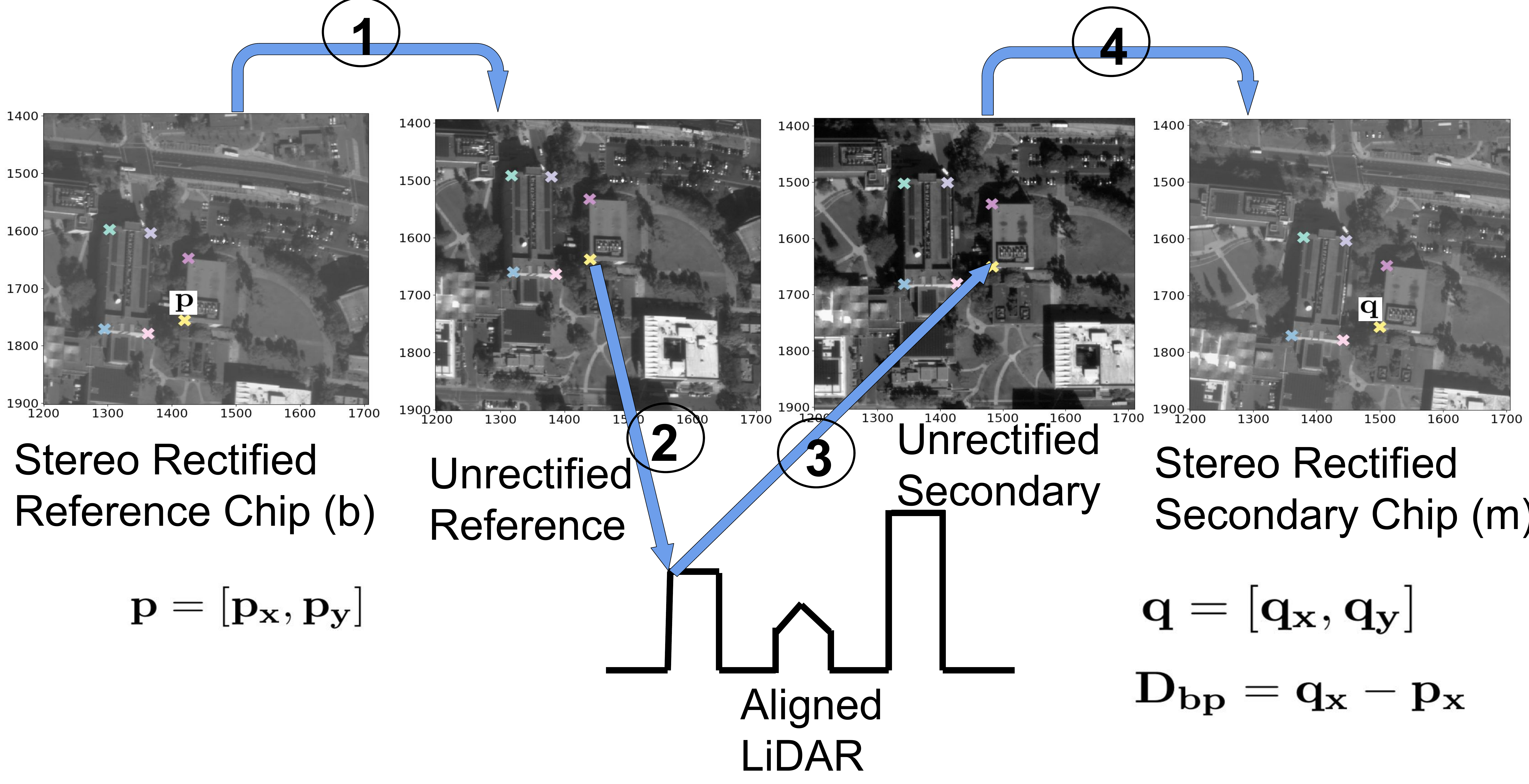}
\end{center}
\caption{Overview of our groundtruthing process. Referring
  to labels 1-4 in the figure, (1) We map points from stereo
  rectified reference view to unrectified reference view (2)
  we backproject these point into aligned LiDAR using
  inverse RPC, giving us latitude, longitude and height for
  each point (3) we forward project these world points onto
  secondary unrectified view (4) Using inverse rectification
  map we map these points into stereo rectified secondary
  view. After performing steps 1 through 4, we can now apply
  the definition of disparity to get disparity value at each
  point.  }
\label{fig:GT_Disp}
\end{figure*}
%-------------------------------------------------------------------------
\subsection{Stereo Rectification}
\label{subsec:stereo_rect}
In this section we will go over the details of chip
rectification. First, we will cover the details of
tile-based rectification in Section
\ref{subsec:tile_stereo_rect}. Then, in Section
\ref{subsec:chip_stereo_rect} we will cover the details of
how we use the output from the tile-based rectification step to
stereo rectify the full chips.
\subsubsection{Stereo Rectification --- Tile Based}
\label{subsec:tile_stereo_rect}
For stereo rectifying tiles, we have used the approach
proposed by \cite{de2014automatic}. We first approximate an RPC
projection function into an affine projection with first order
Taylor series approximation. Let $\mathcal{P}_{RPC} :
\mathbb{R}^3 \rightarrow \mathbb{R}^2$ be the RPC projection
function. The first order Taylor series expansion of
$\mathcal{P}_{RPC}$ around a 3D world point $\bf{X}_o$ can
be given as

\begin{align*}
\mathcal{P}_{RPC}(\bf{X}) &= \mathcal{P}_{RPC}(\bf{X}_o)+ \nabla \mathcal{P}_{RPC}(\bf{X}_o) (\bf{X}-\bf{X}_o) \\
&= \nabla \mathcal{P}_{RPC}(\bf{X}_o)\bf{X} + \bf{b}
\end{align*}

\noindent where $\bf{b} = \mathcal{P}_{RPC}(\bf{X}_o) -
\nabla \mathcal{P}_{RPC}(\bf{X}_o)\bf{X}_o$ and $\nabla
\mathcal{P}_{RPC} (\centerdot)$ is the Jacobian matrix. The
affine approximation of $\mathcal{P}_{RPC} (\centerdot)$ can
be expressed as a $3 \times 4$ matrix in homogeneous
coordinates as follows

$$\mathcal{P}_{Affine}(\bf{X}) = \begin{bmatrix} \nabla \mathcal{P}_{RPC}(\bf{X}_o)  &  \bf{b} \\ 
								0 & 1
\end{bmatrix}^{3 \times 4} \begin{bmatrix} \bf{X} \\ 1 \end{bmatrix}^{4 \times 1}$$

After this step, we can apply off-the-shelf algorithms for
stereo rectifying each pair. For the sake of completeness we
will summarize those steps here. First, we find the
correspondences $\bf{x}_j \leftrightarrow \bf{x}'_{j}$ using
SIFT matches (\cite{lowe2004distinctive},
\cite{otero2015anatomy}), where $\bf{x}_j$ is a point in a
reference view and $\bf{x}'_j$ is the corresponding point in
the secondary view. Then we estimate the fundamental matrix
$F$ \cite{hartley2003multiple} with RANSAC, using
$\bf{x}'^{T}_j F \bf{x}_j = 0$. Finally we estimate
resampling homographies $H$ and $H'$ from $F$
\cite{loop1999computing} by solving the following equation
$$F = H'^{T} \begin{bmatrix}
0 & 0 & 0 \\
0 & 0 & -1 \\
0 & 1 & 0 
\end{bmatrix} H
$$  

For stereo reconstruction, these homographies are further
modified by apply required translation so that the tiles
origin is moved to (0,0). We store both $H$ and $H'$ before
applying translation for stereo rectifying full chips as
explained in the next section \ref{subsec:chip_stereo_rect}.

%%%%%%%%%%%%%%%%%%%%%%%%
%%%%%%%%%%%%%%%%%%%
\begin{figure*}
  \begin{center}    
%\fbox{\rule{0pt}{1.5in} \rule{0.9\linewidth}{0pt}}
\includegraphics[width=\textwidth]{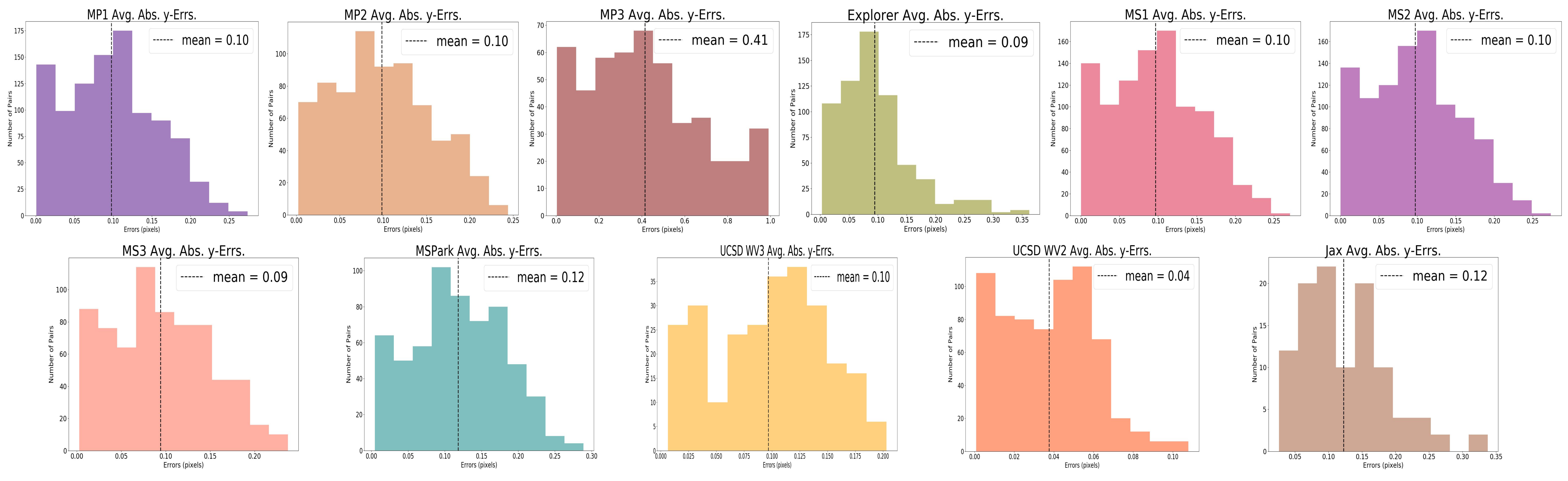}
\end{center}
   \caption{Average chip-level rectification error (absolute
     y-error) distributions for all eleven datasets.}
\label{fig:rect_errs}
\end{figure*}
%%%%%%%%%%%%%%%%%%%
%%%%%%%%%%%%%%%%%%%%%%%%%

\subsubsection{Stereo Rectification --- Chip Based}
\label{subsec:chip_stereo_rect}

Given two images $l$ and $r$, and a point $\bf{p}_l$ in
image $l$ then we can define the parameteric equation for
the epipolar curve as follows:

$epi^{\bf{p}_l}_{lr}(h) = \mathcal{P}_{RPC_r} (\mathcal{P}^{-1}_{RPC_l}(\bf{p}_x, \bf{p}_y, h)) $

Intuitively, it contains the locations for all possible
correspondences in the secondary image for a given point
$\bf{p}_l$ in the reference image for different height
values for $h$. For a pinhole camera, epipolar lines form
conjugate pairs. However, this property does not hold true
for the epipolar curves for the pushbroom cameras. This
makes the stereo rectification problem more challenging for
satellite images.

%% \magenta{For the case of pinhole camera model, for a point
%%   $\bf{p}$ in any one of the two views, the corresponding
%%   point, if it exists, lies on the epipolar line
%%   $epi^{\bf{p}}$ of $\bf{p}$. Therefore, the epipolar lines
%%   form a conjugate pair.

For rectifying full chips we take piecewise homographies and
stitch them together. Subsequently, we compute x- and y-maps
that can transform a grid in the stereo rectified coordinate
space to the unrectified image coordinate space. Since
unrectified chip pairs are broken into non-overlapping
tiles, simply stitching the corresponding rectified tiles
can result in missing image information near the edges.
%as shown in Figure \ref{fig:stereo_rect}.
In order to get smoother boundaries, we use overlapping
tiles and average the x- and y- coordinates in the
overlapping regions in the rectified space. We generate both
rectification and inverse rectification maps in this step,
in order to go back and forth between an unrectified view
and the corresponding stereo rectified view.

% In order to get smoother boundaries, we apply some padding
% around each tile i.e., convert non-overlapping tiles into
% overlapping tiles, and once they are mapped into stereo
% rectified space, we average the x- and y- coordinates in the
% overlapping regions. We generate both rectification and
% inverse rectification maps in this step, in order to go back
% and forth between an unrectified view and the corresponding
% stereo rectified view.

%\subsection{LiDAR Alignment}
\subsection{Generating a Chip-Level Fused DSM}
\label{subsec:lidar_align}

%Once we have generated a set of stereo pairs, we feed them
%to a multi-view stereo reconstruction pipeline. For
%generating fused DSM we have extended s2p's code
%(\cite{de2014automatic}, \cite{facciolo2017automatic}) to
%generate a fused DSM. The publicly available version of s2p
%\cite{s2p_code} supports only pairwise DSM generation.

We now briefly describe our procedure to obtain chip-level
pairwise DSMs and a fused DSM for each AOI. %  We have
% extended the publicly available version of the s2p pipeline
% in order to generate the fused chip-level DSMs that are
% called for by the box labeled ``Fuse Pairwise Chip DSMs'' in
% Figs. 2 and 3.  The publicly available version of the
% pipeline only provides for the generation of pairwise DSMs.

%s2p divides each chip pair into tiles and perform tile based stereo
%reconstruction. 
Using water masks from SRTM DEM, the water regions in the
individual tiles are masked out and such points are marked
invalid in the output DSM.  As explained in Section
\ref{subsec:tile_stereo_rect} each tile pair is stereo
rectified and then we use SGM \cite{hirschmuller2008stereo}
for stereo matching. Then using the estimated tile-level
disparity maps we perform triangulation to get a point cloud
per tile. All the pairwise tile-level pointclouds are merged
to form a pairwise chip-level pointcloud which is converted
into a pairwise chip-level DSM.

After obtaining a sufficient number of pairwise chip-level
DSMs, we fuse these by taking the median over all valid
height values at each point in the output grid. We then use
the alignment tool provided by \cite{pubgeo} for estimating
the required translation to align LiDAR to the fused DSM.

\section{Generating the Disparity Groundtruth from LiDAR}
\label{sec:generate_GT}

%In this section we will go over the details of the method used for
%stereo-rectification of the full chips and generating groundtruth
%disparity maps.

%\subsection{Dataset Generation}
%\label{subsec:dataset_gen}

Figure \ref{fig:GT_Disp} shows the overview of our process
for generating groundtruth disparity maps. We first take all
the points from a stereo rectified reference view and map
into the corresponding unrectified reference view. This is
shown as the arrow with Label 1 in Figure
\ref{fig:GT_Disp}. Then we backproject these point
coordinates, onto LiDAR, using the corresponding RPC
model. Note that the backprojected ray may intersect the
LiDAR at multiple points, so we ensure that the system
returns the point that is actually visible from the
satellite, i.e. the point with the greatest height.
%% One challenge during this step is that, if we do a
%% naive binary search as done in gdal library
%% \magenta{\bf{TODO: add reference}} it can return any
%% intersection point along the backprojected ray instead of
%% the first intersection point. Therefore, we have implemented
%% a raytracing-like approach to get first intersection points
%% along all the backprojected rays from the pixel coordinates
%% in a unrectified view.
We use bilinear interpolation for missing points in LiDAR
e.g. building walls. This step returns a world point
(latitude , longitude and height) for each backprojected
image point. This step is marked by the arrow with Label 2
in Figure \ref{fig:GT_Disp}. In step 3 we take all these
world points and project them onto the unrectified secondary
view using its RPC model. In the last step 4, we map these
point into the secondary rectified view. After these steps 1
through 4 we get correspondences in two stereo rectified
views, as shown by labeled points $\bf{p}$ and $\bf{q}$ in
Figure \ref{fig:GT_Disp}. Then we can use the definition of
disparity to compute a reference disparity map
$\textbf{D}_{b}$. As shown in Figure \ref{fig:GT_Disp},
disparity at point $\bf{p}$ can be calculated
as $$\textbf{D}_{b\bf{p}} = \bf{q}_x-\bf{p}_x$$  
%\textcolor{red}{\Large\bf  Sonali, please fix convention.} 
  We repeat the same process
by switching the order of the two views to get the
corresponding secondary disparity map $\bf{D}_{m}$. Then we
perform a consistency check in the form of the
Left-Right-Right-Left (LRRL) check to detect and mark
occluded pixels as invalid, which is given as

\begin{equation}
%\begin{eqnarray}
  \textbf{D}_{b \textbf{p}} = \begin{dcases*} 
        \textbf{D}_{b \textbf{p} }  &   if $ | \textbf{D}_{b\textbf{p}} - \textbf{D}_{m \textbf{s}} | \leq 1 $ \\
        invalid & otherwise
        \end{dcases*} 
%\text{where} \quad  \textbf{s} =  [ \textbf{p}_x+\textbf{D}_\textbf{p}, \textbf{p}_y]^T 
% \end{eqnarray}
 \label{eq:lrrl}
%\end{subequations}
\end{equation}
where $\textbf{s} =  [ \textbf{p}_x+\textbf{D}_\textbf{p}, \textbf{p}_y]^T $

\subsection{Creating Building Masks}
\label{subsec:building_masks}
For generating building masks for the rectified chips, we
obtain an initial building mask in lat/long space by
applying the tool from \cite{pubgeo} on the aligned
LiDAR. However, we noticed that occasionally, some trees or
vegetation do get marked as buildings. Therefore, we
manually clean up the initial masks. Then we project the
points corresponding to buildings onto unrectified
chips. Finally using the inverse rectification maps, we map
these masks into the rectified chips.
\section{Results}
\label{sec:results}
This section is organized as follows. We first present a
quantitative evaluation of our rectification errors in
Section \ref{subsec:rect_errs} and in the following section
\ref{subsec:disp_errs} we present a quantitative evaluation
of our groundtruth disparity maps using human annotated tie
points. Figure \ref{fig:qualitative_results} shows some
example images and groundtruth disparity maps from the top
four largest AOIs, along with building masks and metadata.

\subsection{Rectification Errors}
\label{subsec:rect_errs}
Since one of the major challenges with using pushbroom
camera models is stereo rectifying the full chips, we
present an evaluation of our stereo rectification method
here. Figure \ref{fig:rect_errs} shows the distribution of
average y-errors across all the pairs in each dataset. For
calculating these errors, we project world points sampled
from a 3D grid onto the unrectified views and then using
inverse rectification maps, we map them into the rectified
views. We then calculate the absolute y-error for each point
and then compute the average. We use around 4000 world
points.  Across all the pairs in each dataset, our average
rectification errors remain within half a pixel.  As can be
seen by the mean values that are displayed separately for
each histogram in Figure \ref{fig:rect_errs}, these errors
are comparable to those for the Middlebury2014 dataset
\cite{Middlebury2014}, for which the reported average error
is 0.2 pixels.

\subsection{Quantitative Evaluation}
\label{subsec:disp_errs}
\begin{figure}[h]
  \begin{center}    
  \includegraphics[width=\linewidth]{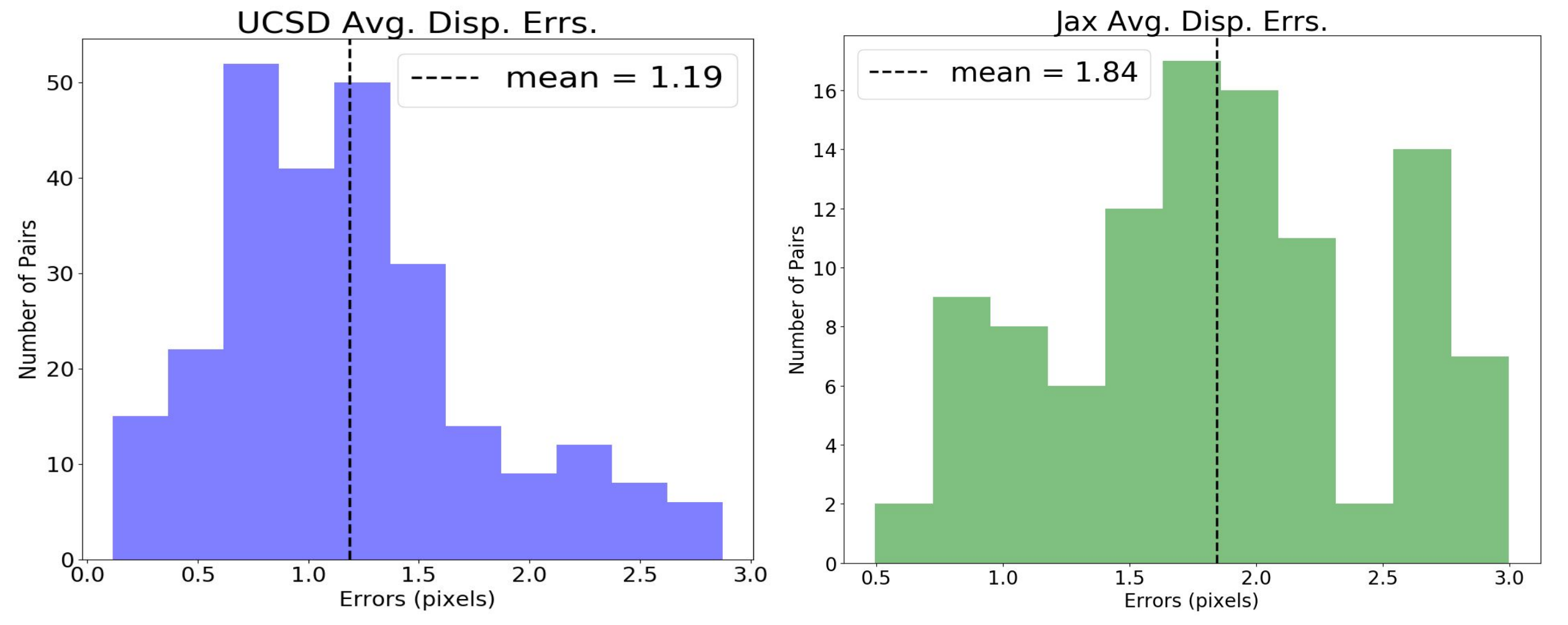}
%\fbox{\rule{0pt}{1.5in} \rule{0.9\linewidth}{0pt}}
\end{center}
   \caption{Disparity errors (using human annotated ground truth points) distribution for UCSD WV3 and Jacksonville AOIs}
\label{fig:disp_errs}
\end{figure}

In this Section we present a quantitative evaluation of the
disparity maps for the two largest AOIs - UCSD and
Jacksonville. We have collected some human annotated tie
points in some views and we use them to quantitatively
evaluate the errors in our groundtruthed disparity
maps. Figure \ref{fig:disp_errs} shows disparity error
distribution over all the groundtruth disparity pairs of
UCSD WV3 and Jacksonville datasets.  The average disparity
error in UCSD is 1.23 pixels and for Jacksonville it is 1.84
pixels.  These errors are obviously not sub-pixel ---
possibly on account of the fact that the LiDAR values are
only known with 30 cm resolution.

%To the best of our knowledge, none of the existing dataset
%reports disparity errors calculated against human annotated
%ground truth points. Therefore, without any reference point
%it's difficult to conclude if we're doing better or worse
%than the existing benchmarking datasets.

\subsection{Stereo Matching Experiment}
\label{subsec:stereo_exp}
\begin{figure}[h]
  \begin{center}    
%\fbox{\rule{0pt}{1.5in} \rule{0.9\linewidth}{0pt}}
\includegraphics[width=\linewidth]{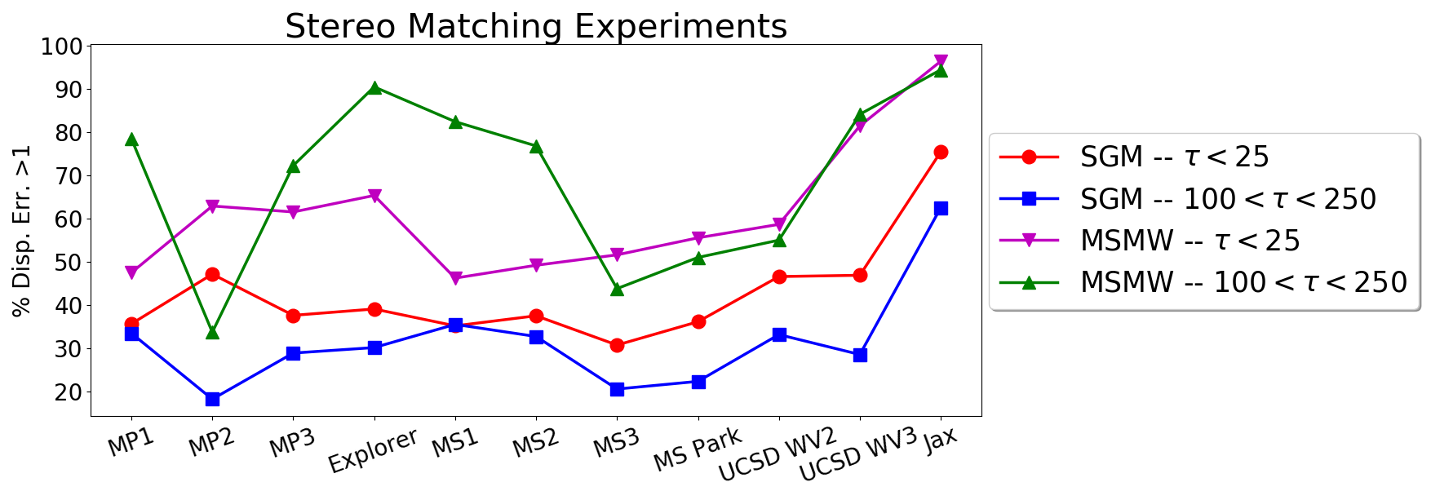}
\end{center}
\caption{% Stereo Matching results , the points labeled as
  % ``SGM -- $\tau < 25$" are results using SGM on a pair,
  % where the time interval is less than a month, similarly
  % the points labeled ``SGM -- $ 100< \tau < 150$" are
  % results using SGM on a pair where the interval is between
  % 100 days and 250 days.
  Stereo Matching results -- ``$\tau < 25$" denotes results
  on pairs where the time interval is less than a month and
  ``$ 100< \tau < 150$" denotes results on pairs where the
  interval is between 100 to 250 days.}
\label{fig:stereo_results}
\end{figure}

In this section we show stereo matching results to
illustrate the challenges posed by out-of-date stereo
pairs. Using the groundtruthed disparities in our datasets,
Figure \ref{fig:stereo_results} shows the percentage of
pixels where the errors exceed one pixel in the estimated
disparties using the SGM \cite{hirschmuller2008stereo} and
MSMW \cite{buades2015reliable} algorithms.  We also show two cases with
regard to the interval between the image acquisition times.
% Using the metadata provided by our dataset, 
In one case the time interval is less than one month and in
the other case it is between 100 days and 250 days. We use the building masks to evaluate the errors. 

\section{Conclusion}
\label{sec:conclusion}
We have contributed a large benchmarking stereo dataset for
out-of-date satellite images and also provided a framework
for how such a dataset can be constructed.  In the dataset
we make available, the rectification accuracy is comparable
to the existing state-of-the-art datasets. Unlike the
existing benckmarking datasets, we have also carried out a
quantitative evaluation of our groundtruthed disparities
using human annotated points in two AOIs. Our stereo
matching experiments show that this dataset presents a new
level of challenge for stereo matching algorithms, both in
terms of stereo-pair sizes and scene variations. We hope
that researchers in the stereo reconstruction and remote
sensing areas will benefit from this dataset.

\section{Acknowledgements}
Supported by the Intelligence Advanced Research Projects
Activity (IARPA) via Department of Interior / Interior
Business Center (DOI/IBC) contract number D17PC00280. The
U.S. Government is authorized to reproduce and distribute
reprints for Governmental purposes notwithstanding any
copyright annotation thereon. Disclaimer: The views and
conclusions contained herein are those of the authors and
should not be interpreted as necessarily representing the
official policies or endorsements, either expressed or
implied, of IARPA, DOI/IBC, or the U.S. Government.

{\small
\bibliographystyle{ieee}
\bibliography{egbib}

\begin{thebibliography}{10}\itemsep=-1pt

\bibitem{pubgeo}
Open source geospatial tools for 3d registration and scene classification.
\newblock
  \url{https://www.jhuapl.edu/pubgeo/170807-FOSS4G-JHUAPL-Open-Source-Geospatial-Tools.pdf}.

\bibitem{TOA_WV2}
Radiometric use of worldview-2 imagery.
\newblock
  \url{https://dg-cms-uploads-production.s3.amazonaws.com/uploads/document/file/104/Radiometric_Use_of_WorldView-2_Imagery.pdf}.

\bibitem{TOA_WV3}
Radiometric use of worldview-3 imagery.
\newblock \url{
  https://dg-cms-uploads-production.s3.amazonaws.com/uploads/document/file/207/Radiometric_Use_of_WorldView-3_v2.pdf}.

\bibitem{Triggs2000}
{Bundle Adjustment-A Modern Synthesis}.
\newblock {\em Vision Algorithms}, 34099:298--372, 2000.

\bibitem{Bay2006}
{SURF: Speeded up robust features}.
\newblock volume 3951 LNCS, pages 404--417, 2006.

\bibitem{U3D}
M.~{Bosch}, K.~{Foster}, G.~{Christie}, S.~{Wang}, G.~D. {Hager}, and
  M.~{Brown}.
\newblock Semantic stereo for incidental satellite images.
\newblock In {\em 2019 IEEE Winter Conference on Applications of Computer
  Vision (WACV)}, pages 1524--1532, Jan 2019.

\bibitem{MVS_Challenge2016}
M.~Bosch, Z.~Kurtz, S.~Hagstrom, and M.~Brown.
\newblock A multiple view stereo benchmark for satellite imagery.
\newblock In {\em 2016 IEEE Applied Imagery Pattern Recognition Workshop
  (AIPR)}, pages 1--9. IEEE, 2016.

\bibitem{Core3D_Public}
M.~Brown, H.~Goldberg, K.~Foster, A.~Leichtman, S.~Wang, S.~Hagstrom, M.~Bosch,
  and S.~Almes.
\newblock Large-scale public lidar and satellite image data set for urban
  semantic labeling.
\newblock In {\em Laser Radar Technology and Applications XXIII}, volume 10636,
  page 106360P. International Society for Optics and Photonics, 2018.

\bibitem{buades2015reliable}
A.~Buades and G.~Facciolo.
\newblock Reliable multiscale and multiwindow stereo matching.
\newblock {\em SIAM Journal on Imaging Sciences}, 8(2):888--915, 2015.

\bibitem{Butler:ECCV:2012}
D.~J. Butler, J.~Wulff, G.~B. Stanley, and M.~J. Black.
\newblock A naturalistic open source movie for optical flow evaluation.
\newblock In {A. Fitzgibbon et al. (Eds.)}, editor, {\em European Conf. on
  Computer Vision (ECCV)}, Part IV, LNCS 7577, pages 611--625. Springer-Verlag,
  Oct. 2012.

\bibitem{de2014automatic}
C.~De~Franchis, E.~Meinhardt-Llopis, J.~Michel, J.-M. Morel, and G.~Facciolo.
\newblock An automatic and modular stereo pipeline for pushbroom images.
\newblock In {\em ISPRS Annals of the Photogrammetry, Remote Sensing and
  Spatial Information Sciences}, 2014.

\bibitem{facciolo2017automatic}
G.~Facciolo, C.~De~Franchis, and E.~Meinhardt-Llopis.
\newblock Automatic 3d reconstruction from multi-date satellite images.
\newblock In {\em Proceedings of the IEEE Conference on Computer Vision and
  Pattern Recognition Workshops}, pages 57--66, 2017.

\bibitem{KITTI2012}
A.~Geiger, P.~Lenz, and R.~Urtasun.
\newblock Are we ready for autonomous driving? the kitti vision benchmark
  suite.
\newblock In {\em Conference on Computer Vision and Pattern Recognition
  (CVPR)}, 2012.

\bibitem{Grodecki2003}
J.~Grodecki and G.~Dial.
\newblock {Block adjustment of high-resolution satellite images described by
  Rational Polynomials}.
\newblock {\em Photogrammetric Engineering and Remote Sensing}, 69(1):59--68,
  2003.

\bibitem{hartley2003multiple}
R.~Hartley and A.~Zisserman.
\newblock {\em Multiple view geometry in computer vision}.
\newblock Cambridge university press, 2003.

\bibitem{hirschmuller2008stereo}
H.~Hirschmuller.
\newblock Stereo processing by semiglobal matching and mutual information.
\newblock {\em IEEE Transactions on pattern analysis and machine intelligence},
  30(2):328--341, 2008.

\bibitem{Middlebury5or6}
H.~Hirschmuller and D.~Scharstein.
\newblock Evaluation of cost functions for stereo matching.
\newblock In {\em 2007 IEEE Conference on Computer Vision and Pattern
  Recognition}, pages 1--8. IEEE, 2007.

\bibitem{loop1999computing}
C.~Loop and Z.~Zhang.
\newblock Computing rectifying homographies for stereo vision.
\newblock In {\em Proceedings. 1999 IEEE Computer Society Conference on
  Computer Vision and Pattern Recognition (Cat. No PR00149)}, volume~1, pages
  125--131. IEEE, 1999.

\bibitem{lowe2004distinctive}
D.~G. Lowe.
\newblock Distinctive image features from scale-invariant keypoints.
\newblock {\em International journal of computer vision}, 60(2):91--110, 2004.

\bibitem{KITTI2015}
M.~Menze and A.~Geiger.
\newblock Object scene flow for autonomous vehicles.
\newblock In {\em Conference on Computer Vision and Pattern Recognition
  (CVPR)}, 2015.

\bibitem{nakamura1996occlusion}
Y.~Nakamura, T.~Matsuura, K.~Satoh, and Y.~Ohta.
\newblock Occlusion detectable stereo-occlusion patterns in camera matrix.
\newblock In {\em Proceedings CVPR IEEE Computer Society Conference on Computer
  Vision and Pattern Recognition}, pages 371--378. IEEE, 1996.

\bibitem{oh2010piecewise}
J.~Oh, W.~H. Lee, C.~K. Toth, D.~A. Grejner-Brzezinska, and C.~Lee.
\newblock A piecewise approach to epipolar resampling of pushbroom satellite
  images based on rpc.
\newblock {\em Photogrammetric Engineering \& Remote Sensing},
  76(12):1353--1363, 2010.

\bibitem{otero2015anatomy}
I.~R. Otero.
\newblock {\em Anatomy of the SIFT Method}.
\newblock PhD thesis, {\'E}cole normale sup{\'e}rieure de Cachan-ENS Cachan,
  2015.

\bibitem{Middlebury2014}
D.~Scharstein, H.~Hirschm{\"u}ller, Y.~Kitajima, G.~Krathwohl,
  N.~Ne{\v{s}}i{\'c}, X.~Wang, and P.~Westling.
\newblock High-resolution stereo datasets with subpixel-accurate ground truth.
\newblock In {\em German conference on pattern recognition}, pages 31--42.
  Springer, 2014.

\bibitem{Middlebury2001}
D.~Scharstein and R.~Szeliski.
\newblock A taxonomy and evaluation of dense two-frame stereo correspondence
  algorithms.
\newblock {\em International journal of computer vision}, 47(1-3):7--42, 2002.

\bibitem{Middlebury2003}
D.~Scharstein and R.~Szeliski.
\newblock High-accuracy stereo depth maps using structured light.
\newblock In {\em 2003 IEEE Computer Society Conference on Computer Vision and
  Pattern Recognition, 2003. Proceedings.}, volume~1, pages I--I. IEEE, 2003.

\bibitem{eth3d}
T.~Schops, J.~L. Schonberger, S.~Galliani, T.~Sattler, K.~Schindler,
  M.~Pollefeys, and A.~Geiger.
\newblock A multi-view stereo benchmark with high-resolution images and
  multi-camera videos.
\newblock In {\em Proceedings of the IEEE Conference on Computer Vision and
  Pattern Recognition}, pages 3260--3269, 2017.

\end{thebibliography}
}

\end{document}